\documentclass[11pt]{article}

\IfFileExists{acl.sty}{
  \usepackage[]{acl}
}{
  \usepackage[margin=1in]{geometry}
  \usepackage[round]{natbib}
}
\usepackage{fvextra}
\makeatletter
\acl@anonymizefalse
\makeatother

\usepackage{xurl}
\usepackage{hyperref}

\newcommand{\agorasimcode}{\href{https://github.com/haalab/AgoraSim}{code package}}
\newcommand{\agorasimvideo}{\href{https://youtu.be/AdT7bEu3sLU}{demonstration video}}

\DefineVerbatimEnvironment{CodeBlock}{Verbatim}{
  fontsize=\footnotesize,
  breaklines=true,
  breakanywhere=true,
  breaksymbolleft={},
  breaksymbolright={},
  tabsize=2
}
\usepackage{times}
\usepackage{latexsym}
\usepackage[T1]{fontenc}
\usepackage[utf8]{inputenc}
\usepackage{microtype}
\usepackage{inconsolata}
\usepackage{graphicx}
\usepackage{booktabs}
\usepackage{multirow}
\usepackage{xcolor}
\usepackage{xurl}
\usepackage{tikz}
\usetikzlibrary{arrows.meta,positioning,fit,backgrounds}

\title{AgoraSim: A Hybrid Agent-Based Modeling Framework}

\author{
  Chung-Chi Chen \\
  Human-Agent Ally Lab (HAA Lab) \\
  National Institute of Informatics, Japan \\
  \texttt{chen@nii.ac.jp}
}

\begin{document}
\maketitle

\begin{abstract}
LLM-agent simulations make natural-language social scenarios easy to instantiate,
but their outputs can be overread as predictions and are often difficult to
compare with explicit social dynamics. We present AgoraSim, a hybrid
agent-based modeling framework for scenario-oriented social reaction analysis.
AgoraSim resolves textual or multimodal artifacts into editable ABM
configurations, runs ratio-controlled populations that mix LLM, vision-language, custom-endpoint, random, and classical agents, and compares the same scenario
against matched classical reference dynamics. All agents emit a shared structured
decision object, enabling common action spaces, interaction protocols, metrics,
and audit records. Exposed through a local UI, Python SDK/CLI, and REST API,
AgoraSim helps users inspect scenario trajectories, compare modeling assumptions,
and identify cases that warrant empirical validation.
\end{abstract}

\section{Introduction}

Many consequential decisions begin as natural-language or multimodal artifacts:
a policy announcement, a product launch, a public apology, a campaign message,
a social-media post, a video, or an advertisement. Before such an artifact is
deployed, stakeholders often want to explore how different publics might react.
The useful output is rarely a single sentiment label or a point forecast of
public opinion. In early-stage decision making, users need to inspect scenario
trajectories: which reactions are plausible, which social exposures might amplify
or dampen them, which assumptions make the trajectory change, and where a pattern
appears robust or fragile. This is a natural setting for an NLP demo because the
input is language and media, the agents' evidence and rationales are expressed in
language, and the output must remain legible to users comparing alternatives.

We frame this task as \emph{scenario-oriented social simulation}. The goal is
not to reproduce a real population in fine detail or to claim predictive
authority over future behavior. Social simulation is most useful when it helps
explain collective patterns, construct hypotheses, and make modeling assumptions
explicit \citep{wu2026llmsocialsimulation}. This distinction is especially
important for LLM-based social simulation. Language-model agents can produce
fluent and plausible reactions, but plausible text alone does not establish that
a simulation captures behavioral variance, subgroup differences, tipping points,
or path-dependent dynamics. A useful demo should therefore help users see how a
scenario behaves under explicit assumptions, not invite them to treat synthetic
reactions as public opinion.

Agent-based modeling (ABM) provides the classical modeling language for this
objective. ABM represents social phenomena through local rules, heterogeneous
agents, networks, feedback, and interaction over time. Foundational models show
how simple micro-level assumptions can generate macro-level patterns such as
segregation, cascades, adoption, consensus, contagion, and polarization
\citep{schelling1971dynamic,granovetter1978threshold,bass1969newproduct,
degroot1974reaching,clifford1973model,axelrod1997dissemination,
epstein1996growing,bonabeau2002agent,defuant2000mixing}. Its value is not that
any single rule family is universally realistic, but that assumptions are
explicit, parameterized, inspectable, and comparable. ABM gives natural-language
social simulation a methodological scaffold: it turns a scenario into agents,
states, actions, exposure rules, heterogeneity assumptions, and collective
metrics.

Traditional ABM, however, places a heavy modeling burden on the user. To ask how
a community might react to a particular apology, policy brief, advertisement,
image, or product announcement, the analyst must first translate the artifact
into states, thresholds, utilities, networks, and observables. LLM agents invert
this trade-off. They can read rich textual context, respond to open-ended
prompts, produce justifications, and serve as persona-conditioned respondents or
simulated economic and social actors
\citep{argyle2023outofone,aher2023using,horton2023llm}. Interactive systems
such as Social Simulacra, Generative Agents, Concordia, Humanoid Agents, and
LLM-augmented social-network simulators have shown that language-model agents
can populate communities, maintain memory, plan, converse, and exhibit emergent
social behavior
\citep{park2022socialsimulacra,park2023generativeagents,
vezhnevets2023concordia,wang2023humanoidagents,gao2023s3,yang2024oasis}.
More recent systems, including AgentSociety, PolicySim, POSIM, and
discourse-sim, move this line toward large-scale social simulation, platform
intervention, public-opinion evolution, and attitude diffusion
\citep{piao2025agentsociety,huang2026policysim,zhang2026posim,
reji2026attitudediffusion}.

The strengths of LLM agents do not remove the need for ABM; they make ABM more
important. Current LLM agents can collapse toward high-probability or
population-typical responses, sometimes behaving like an ``average persona''
rather than a behaviorally diverse population
\citep{wu2026llmsocialsimulation}. For many social phenomena, the average
reaction is not enough. Polarization, cascades, adoption thresholds, minority
influence, contagion, and path dependence depend on the distribution of
responses and on who influences whom. A demo that only prompts many LLM agents
risks presenting a fluent aggregate without showing whether the trajectory
reflects semantic interpretation, provider bias, prompt sensitivity,
insufficient heterogeneity, or a simple diffusion process that a classical model
would also produce.

This is why comparison with traditional ABM is central to our design. Classical
models are not included merely as historical background or weak baselines. They
serve as explicit reference dynamics. A user can ask whether an LLM-only
population behaves like threshold diffusion, Bass adoption, bounded-confidence
opinion change, contagion, herding, DeGroot/voter social learning, or
discrete-choice utility under the same event, population, network, seed, action
space, and metrics. If an LLM-agent run resembles a simple classical reference,
the observed pattern may not require rich language understanding to explain. If
it diverges, the divergence becomes a scenario hypothesis: the artifact's wording
may matter, the provider may shift reactions, the network may amplify minority
positions, or the scenario may require human-subject validation.

We present AgoraSim, a hybrid agent-based modeling framework for
scenario-oriented social reaction analysis. AgoraSim treats LLM and
vision-language agents as one agent type inside a broader ABM system. A single
run can mix hosted language-model agents, vision-language agents,
OpenAI-compatible local or custom endpoints, random agents, and classical
rule-based agents by ratio. All agents emit the same structured decision object,
so open-ended language behavior is projected into an auditable state space. The
same scenario can also launch matched all-classical reference runs. As a result,
AgoraSim turns an open-ended prompting workflow into a comparable ABM experiment
with shared action spaces, interaction protocols, metrics, reference dynamics,
and records.

The contribution of AgoraSim is threefold. First, it provides a natural-language
scenario workbench that resolves text or media artifacts into editable ABM
configurations rather than opaque prompts. Second, it provides a hybrid
execution substrate where LLM, VLM, custom-endpoint, random, and classical agents
coexist through ratio-controlled model slots and shared structured outputs.
Third, it provides an analyst-facing demo artifact---local UI, Python SDK/CLI,
REST API, saved configurations, cost accounting, and per-agent audit records---
for comparing scenario trajectories under explicit modeling assumptions. The
intended use is exploratory and comparative: agreements between LLM agents and
classical dynamics can reveal qualitative patterns that are robust across
assumptions, while divergences expose which modeling choices matter and which
cases deserve empirical validation.

\section{System Design}
\label{sec:system}

\begin{figure*}[t]
    \centering
    \includegraphics[width=0.98\linewidth]{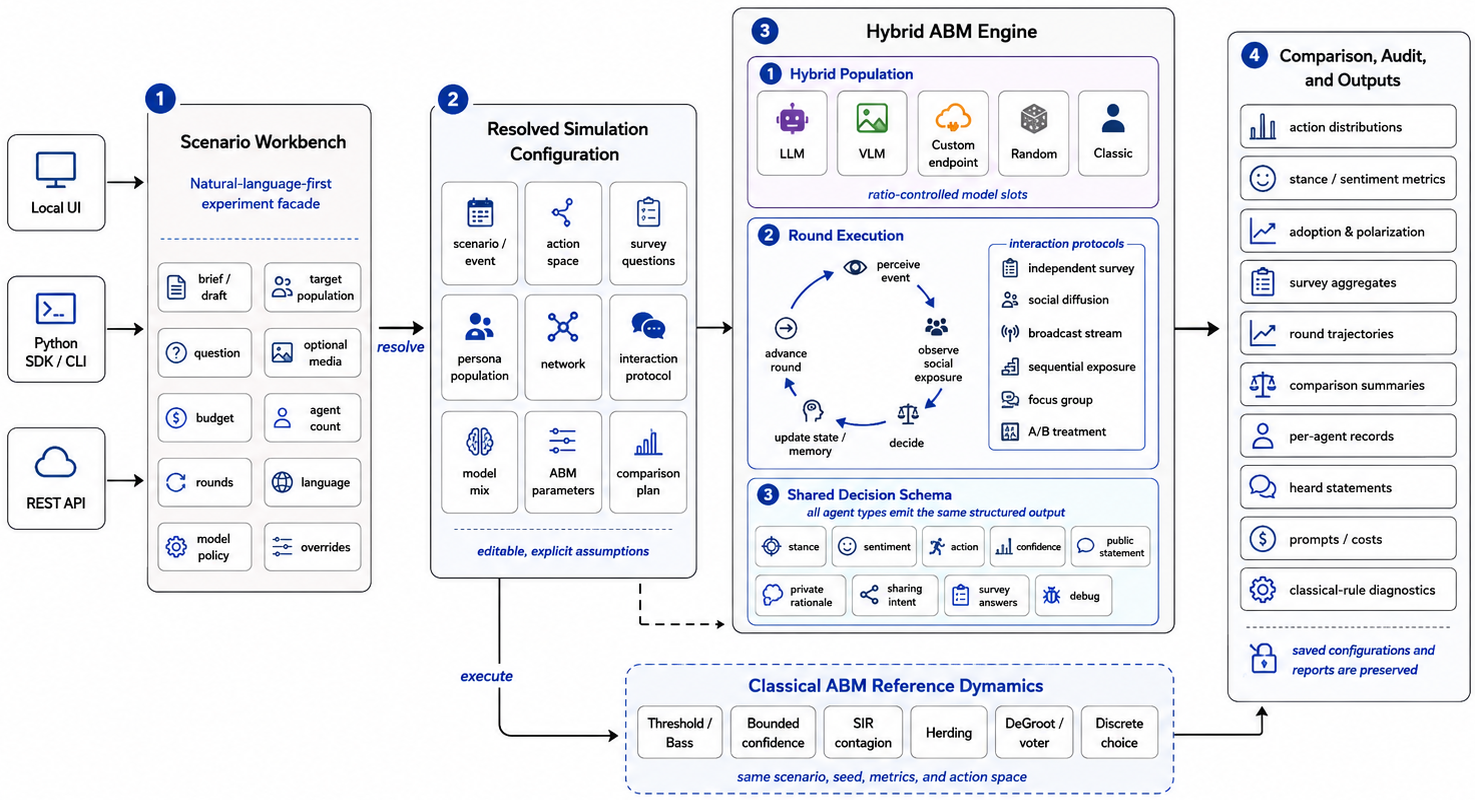}
    \caption{AgoraSim architecture. A natural-language or multimodal artifact is
    resolved into an editable simulation configuration, executed by a
    ratio-controlled hybrid ABM engine, and compared with matched classical
    reference dynamics through shared metrics and audit records.}
    \label{fig:architecture}
\end{figure*}

Figure~\ref{fig:architecture} summarizes the system. The design principle behind
AgoraSim is that a natural-language scenario should not remain an opaque prompt:
it should become an explicit ABM configuration whose assumptions can be
inspected, edited, executed, and compared. The system therefore separates three
concerns. The scenario workbench turns language and media inputs into structured
simulation settings. The hybrid ABM engine executes LLM, VLM, custom-endpoint,
random, and classical agents under a shared action space and interaction
protocol. The comparison and audit layer runs matched classical reference
dynamics and stores the records needed to understand where a trajectory came
from.

\subsection{From Artifacts to Simulation Configurations}

A run begins from a compact user-facing description: a brief or content draft,
target population, analysis question, optional media, budget, agent count,
rounds, language, model policy, and optional overrides. The resolver expands
these inputs into an editable simulation configuration containing the event,
persona population, action space, survey questions, network, interaction
protocol, model mix, ABM parameters, and comparison plan. The composer lowers the
cost of scenario construction, but it does not hide the model: before execution,
the run is materialized as a configuration that the user can inspect and revise.

The action space is the key bridge between NLP and ABM. LLM agents can produce
free-form statements and rationales, but a simulation also needs stable variables
that can be counted, compared, fed into later rounds, and plotted against
classical dynamics. AgoraSim therefore represents each scenario with a compact set
of mutually interpretable actions, such as \textsc{support}, \textsc{oppose},
\textsc{share}, \textsc{ignore}, \textsc{wait}, or domain-specific alternatives.
This keeps natural-language explanations available while giving all agent types
a common state representation.

For multimodal scenarios, the configuration records how media is perceived. If a
selected provider supports vision, agents may receive the artifact directly. For
agents without vision capability, AgoraSim can generate a neutral caption once
and reuse it across calls. This avoids turning media perception into an
uncontrolled source of variation and keeps the perceived artifact visible in the
audit trail.

\begin{table*}[t]
\centering
\small
\begin{tabular}{p{0.24\linewidth}p{0.68\linewidth}}
\toprule
Reference dynamic & What it helps users inspect \\
\midrule
Threshold / Bass &
Whether adoption can be explained by thresholds, independent uptake, and
imitation pressure. \\
Bounded confidence / Deffuant &
Whether opinion movement is constrained by similarity and whether polarization
persists. \\
SIR contagion &
Whether attention, rumor spread, or alarm follows exposure-driven spread and
decay. \\
Herding &
Whether crowd pressure dominates independent re-evaluation. \\
DeGroot / voter &
Whether consensus or majority switching explains the observed trajectory. \\
Discrete choice &
Whether decisions follow interpretable utilities such as trust, risk, price
sensitivity, habit, or social proof. \\
\bottomrule
\end{tabular}
\caption{Classical ABM reference dynamics supported by AgoraSim. Each can be used
as an in-population agent type or as a matched comparison run under the same
scenario and metric schema.}
\label{tab:strategies}
\end{table*}

\subsection{Hybrid Agents and Round Execution}

The central execution object is the model mix. Each slot specifies a provider,
model, weight, temperature, media capability, and optional strategy parameters.
Slots may refer to hosted LLMs, vision-language models, local or custom
OpenAI-compatible endpoints, random agents, or the \texttt{classic} provider.
At runtime, agents are sampled from the target population and assigned to slots
according to user-controlled ratios. Thus AgoraSim can run LLM-only,
classical-only, random-only, or mixed populations without changing the scenario
definition.

All agent types implement the same decision interface. Given the current round
context, an agent returns a structured object containing stance, sentiment,
selected action, confidence, public statement, private rationale, sharing intent,
survey answers, and debug fields. LLM and VLM agents produce this object through
prompts grounded in persona, event context, optional media, memory, and social
exposure. Classical agents update internal states through the selected rule
family and map those states to the same action space. Random agents provide a
sanity-check population. This shared schema prevents LLM behavior from remaining
only free-form text and prevents classical rules from living in a separate
numeric simulator.

\begin{figure*}[t]
\centering
\begin{minipage}[t]{0.49\linewidth}
\centering
\includegraphics[width=\linewidth,trim=245 0 0 0,clip]{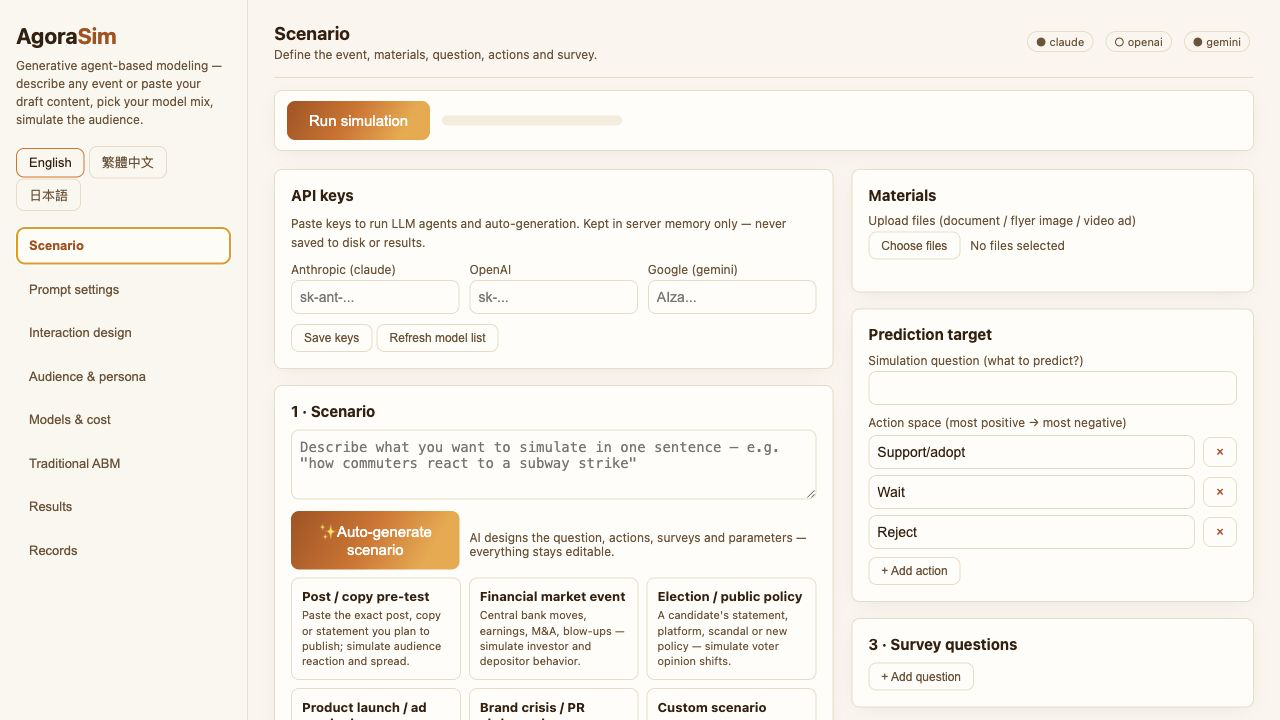}
\vspace{2pt}
{\small (a) Scenario setup}
\end{minipage}
\hfill
\begin{minipage}[t]{0.49\linewidth}
\centering
\includegraphics[width=\linewidth,trim=245 0 0 0,clip]{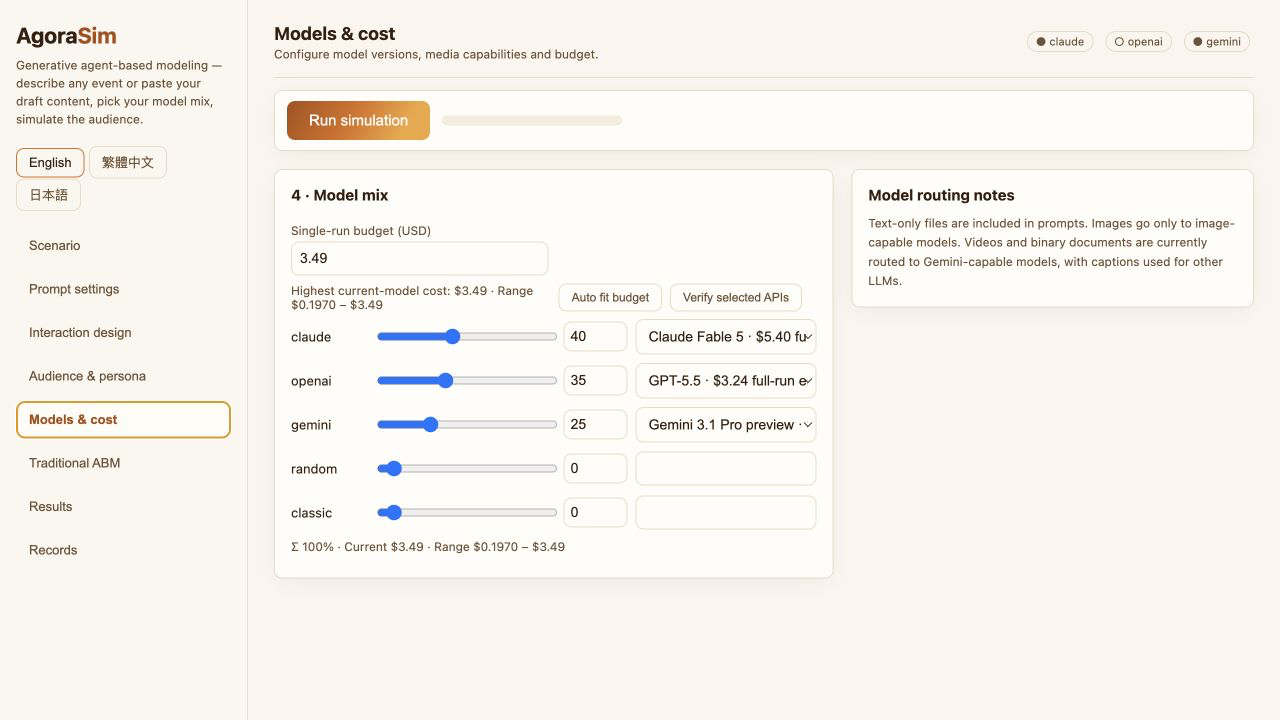}
\vspace{2pt}
{\small (b) Model mix}
\end{minipage}
\caption{Creating a scenario experiment. Users enter a natural-language or
multimodal artifact, edit the generated action space and survey fields, and
choose a ratio-controlled mix of LLM, VLM, custom, random, and classical agents.}
\label{fig:workflow_setup}
\end{figure*}

\begin{figure*}[t]
\centering
\begin{minipage}[t]{0.49\linewidth}
\centering
\includegraphics[width=\linewidth,trim=245 0 0 0,clip]{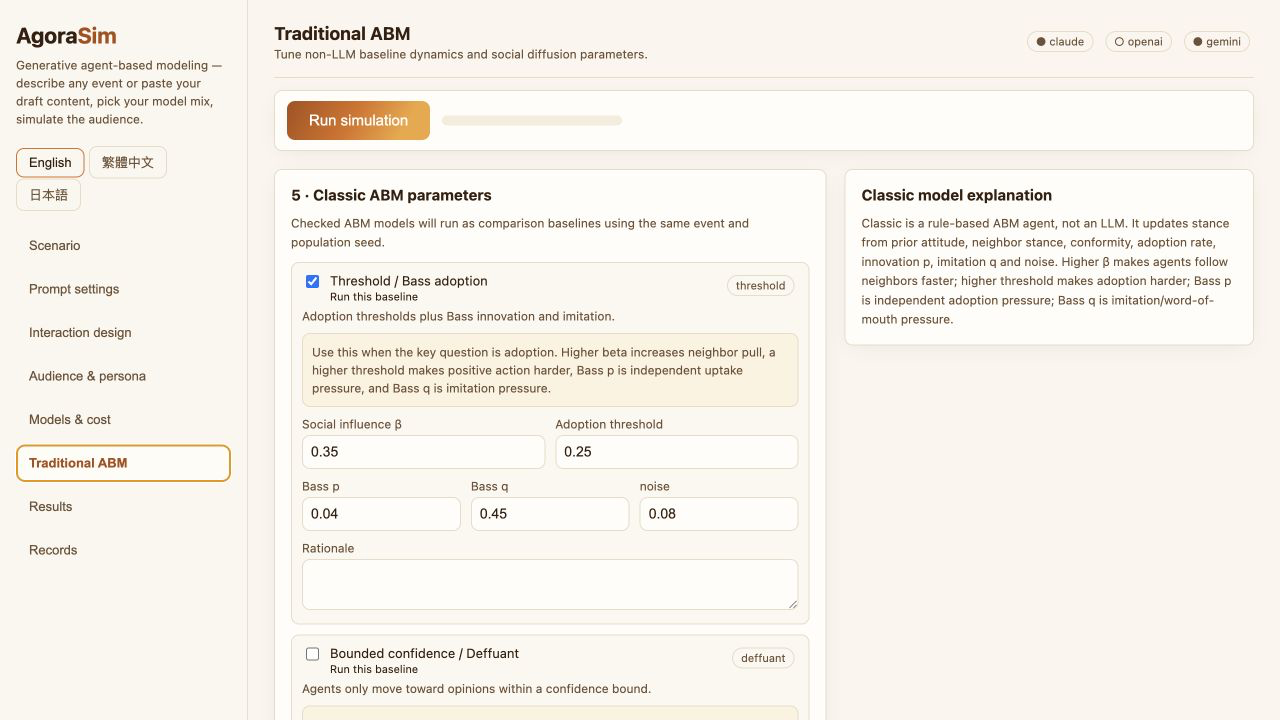}
\vspace{2pt}
{\small (a) Classical reference dynamics}
\end{minipage}
\hfill
\begin{minipage}[t]{0.49\linewidth}
\centering
\includegraphics[width=\linewidth,trim=245 0 0 0,clip]{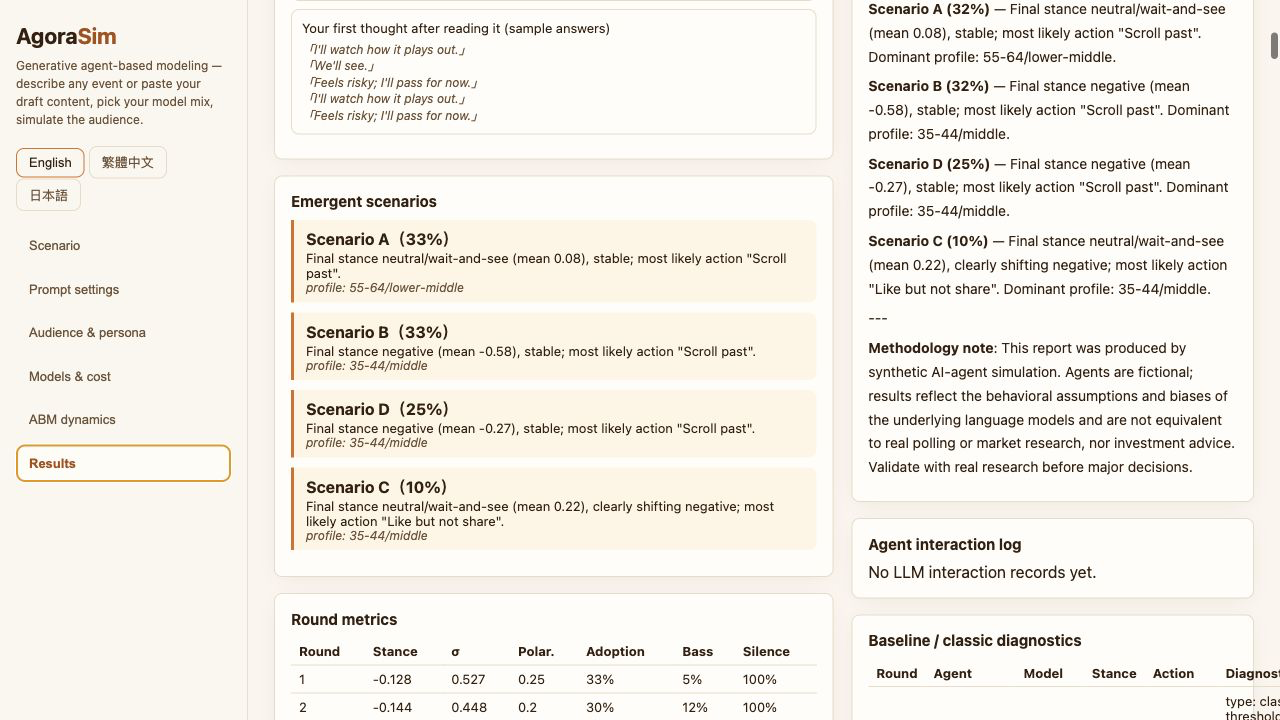}
\vspace{2pt}
{\small (b) Results and audit view}
\end{minipage}
\caption{Comparison and inspection. Users select classical ABM reference
dynamics with editable parameters, then inspect emergent scenario summaries,
round metrics, methodology notes, and diagnostic records after execution.}
\label{fig:workflow_results}
\end{figure*}

Runs proceed over discrete rounds. During initialization, the engine samples
personas, assigns model slots, builds the network, resolves media handling, and
prepares the comparison bundle. In round \(t\), each agent receives the event,
persona, action space, survey questions, optional media, prior memory, aggregate
summaries, and interaction inputs derived from round \(t-1\). After all agents
emit decisions, the environment updates aggregate state, stores per-agent memory,
computes metrics, and advances to the next round. The current implementation
supports several visible interaction protocols, including independent survey,
social diffusion through network neighbors, broadcast comment streams,
sequential exposure, focus groups, and A/B treatment. These protocols are kept
simple and editable so that the exposure assumption remains part of the
experiment rather than a hidden platform model.

\subsection{Reference Dynamics, Comparison, and Audit}

Classical ABM enters the system in two ways. First, \texttt{classic} can be used
as an in-population provider, allowing rule-based agents to participate in the
same network and action space as LLM agents. Second, the comparison controller
can launch all-classical reference runs using the same scenario, seed, population
assumptions, network settings, rounds, metrics, and action space. These reference
dynamics are not included merely as weak baselines. They give meaning to an
LLM-agent trajectory: if a hybrid run resembles a simple threshold or contagion
model, the pattern may be explainable by exposure dynamics; if it diverges, the
divergence becomes a scenario hypothesis to inspect. Table~\ref{tab:strategies}
summarizes the reference dynamics currently supported by AgoraSim.

The comparison bundle launches the current run and selected reference dynamics
together. The current run may be LLM-only, VLM-enabled, classical-only,
random-only, or hybrid; the reference runs are all-classical configurations using
the selected rule families. Because the runs share the same scenario, action
space, interaction protocol, metrics, and seed when applicable, their qualitative
trajectories can be compared directly.

AgoraSim records both aggregate outputs and the evidence behind them. Aggregate
views include action distributions, stance and sentiment summaries, adoption or
polarization measures, survey aggregates, provider shares, cost summaries, round
trajectories, and comparison summaries. The record layer stores the resolved
configuration, model assignments, prompts, parsed decision objects, public
statements, heard neighbor statements, survey answers, memory, token and latency
metadata when available, degradation status, and classical-rule diagnostics.
This supports drill-down from a macro-level curve to the scenario assumptions,
agent exposures, and decisions that produced it.

The same engine is exposed through a local browser/desktop UI, Python SDK/CLI,
and REST API. All entry points resolve to the same configuration and produce the
same comparison outputs and audit records. Offline random and classical modes
allow reviewers to exercise the artifact without API keys, while live LLM/VLM
runs use user-owned keys and can be compared against the same saved reference
dynamics.

\section{Demo Workflow and Artifact}
\label{sec:workflow}

The demo is organized as a scenario walkthrough rather than an accuracy
evaluation. A user begins with a brief, draft, or media artifact; turns it into
an explicit simulation configuration; chooses a heterogeneous agent population;
selects classical reference dynamics; and then inspects aggregate trajectories
and audit records. The intended path is from an open-ended question, such as how
commuters might react to congestion pricing, to a family of comparable scenario
trajectories.

Figure~\ref{fig:workflow_setup} shows the first half of the workflow. In the
scenario setup view, users can paste a draft announcement, upload optional media,
or enter a short brief. The composer proposes an event description, research
question, action space, survey questions, and related simulation settings, but
these fields remain editable assumptions rather than final model outputs. The
model-mix view then exposes heterogeneity as an experimental variable: users can
assign weights to hosted LLMs, vision-language models, local or custom
OpenAI-compatible endpoints, random agents, and classical rule-based agents.

Figure~\ref{fig:workflow_results} shows the comparison and inspection stages.
The classical ABM view lets users enable reference dynamics such as
threshold/Bass adoption, bounded confidence, contagion, herding, DeGroot/voter
learning, and discrete choice, with editable parameters. Each selected reference
is run under the same scenario, action space, seed, population assumptions,
network settings, and metrics when applicable. The results view then presents
scenario summaries, round metrics, comparison outputs, and diagnostic records so
that users can trace an aggregate pattern back to the assumptions and decisions
that produced it.

\paragraph{Execution surfaces.}
The same workflow is available through a local browser/desktop UI, Python
SDK/CLI, and REST API. The local UI supports interactive use and keeps scenario
drafts under the user's control; live LLM/VLM calls use user-owned API keys,
which are entered locally and not written to result files. The SDK/CLI supports
scripted experiments and sensitivity sweeps over content variants, model mixes,
interaction protocols, seeds, and ABM reference strategies. The REST API exposes
the same compact experiment facade for integration with dashboards, annotation
tools, educational interfaces, or larger research pipelines. All entry points
resolve to the same saved configuration and produce the same comparison outputs
and audit records. The runnable \agorasimcode{} and \agorasimvideo{} are public;
the package includes the local UI, server, SDK/CLI, REST API, examples,
packaging scripts, and documentation, but no API keys.

\section{Example Scenario}
\label{sec:example}

We demonstrate AgoraSim with a congestion-pricing scenario. The user enters a
draft policy announcement and asks how urban commuters might react to a proposed
city fee, whether support changes if transit improvements are promised, and how
reactions spread through discussion. The scenario workbench turns this artifact
into an editable configuration: an event description, a target population of
commuters, a compact action space such as \textsc{support}, \textsc{oppose},
\textsc{wait}, and \textsc{adapt}, and survey questions about fairness,
affordability, trust, and likely travel behavior.

The user then runs the same scenario under multiple assumptions. One hybrid run
uses a social-diffusion protocol over a small-world network and mixes LLM agents
with classical agents. In parallel, AgoraSim launches two classical reference
dynamics. A threshold/Bass reference asks whether support can be explained by
exposure, imitation, and adoption pressure. A discrete-choice reference asks
whether reactions follow interpretable utilities such as cost sensitivity,
transit access, habit, trust, and perceived fairness. Because all runs share the
same action space and metrics, the dashboard can compare the hybrid trajectory
against these reference dynamics directly.

The value of the example is not a claim that AgoraSim predicts real commuter
opinion. It shows how a user can inspect a family of scenario trajectories. If
the hybrid run closely follows the threshold/Bass reference, the observed pattern
may not require rich language understanding to explain. If the hybrid run
diverges from both references, the divergence becomes a hypothesis to inspect:
perhaps the policy wording changes reactions, perhaps a model provider shifts
the population response, perhaps network exposure amplifies opposition, or
perhaps the scenario needs human-subject validation. The records view supports
this interpretation by linking aggregate curves back to agent statements, heard
neighbor comments, selected actions, and classical-rule diagnostics. In this
sense, the demo turns an open-ended artifact into a comparative scenario
analysis: agreements reveal qualitative patterns that are robust across
assumptions, while divergences expose which modeling choices matter.

\section{Conclusion}

We presented AgoraSim, a hybrid agent-based modeling framework for
scenario-oriented social reaction analysis. The system resolves natural-language
or multimodal artifacts into editable ABM configurations, runs ratio-controlled
populations of LLM, VLM, custom-endpoint, random, and classical agents, and
compares the resulting trajectories with matched classical reference dynamics.
By giving all agents a shared structured decision schema, AgoraSim makes
free-form language behavior comparable through common action spaces, interaction
protocols, metrics, and audit records.

The contribution is not a claim of predictive validity. Instead, AgoraSim is a
demo artifact for inspecting how scenarios behave under explicit assumptions.
Classical ABM references show whether a qualitative trajectory resembles
established dynamics; LLM agents provide semantic responses and natural-language
rationales; and the audit layer exposes the records needed to understand where a
result came from. This makes AgoraSim useful for scenario exploration, teaching
natural-language ABM, and constructing baselines for future research on hybrid
LLM-agent social simulation.

\section*{Impact Statement}

AgoraSim is designed for exploratory scenario analysis, not for measuring or
predicting real populations. Its outputs are synthetic trajectories produced
under explicit modeling assumptions, and should not be used as substitutes for
polling, market research, human-subject experiments, policy evidence,
investment advice, or safety-critical decision making. The intended uses are
hypothesis generation, teaching, method comparison, and baseline construction.

Several risks remain. Synthetic personas may reproduce stereotypes, flatten
minority viewpoints, or reflect cultural and political biases from model
providers. Users may also overinterpret small synthetic populations or misuse
the system to test persuasive framing. AgoraSim mitigates these risks by labeling
reports as synthetic, preserving resolved configurations, showing classical ABM
reference dynamics alongside LLM-agent runs, and exposing prompts, heard
statements, costs, rule diagnostics, and agent-level records. These safeguards
make assumptions inspectable, but they do not establish external validity.

Privacy and reproducibility also require care. The local deployment keeps the
application and saved results under the user's control, and API keys are not
written to result files. However, live LLM/VLM runs still send selected prompts
and inputs to the chosen providers, so users should avoid sensitive artifacts
unless they are authorized to use those services. More broadly, current LLM
agents may collapse toward population-typical responses, and the implemented
networks, memory, and exposure protocols are simplified. Claims about precise
proportions, subgroup reactions, minority viewpoints, tipping points, or
individual trajectories therefore require empirical validation beyond the demo.

\bibliography{references}

\clearpage
\appendix

\section{Full Configuration Schema}
\label{app:config}

Before execution, every AgoraSim run is materialized as a resolved simulation
configuration. This configuration is saved with the result bundle so that users
can inspect, rerun, or modify the assumptions behind a trajectory. It also keeps
the UI, SDK/CLI, and REST API aligned: regardless of the entry point, the run is
defined by the same configuration object. Table~\ref{tab:config_schema}
summarizes the core fields.

\begin{table*}[t]
\centering
\small
\begin{tabular}{p{0.22\linewidth}p{0.68\linewidth}}
\toprule
Field & Description \\
\midrule
\texttt{event} &
Natural-language or multimodal artifact perceived by agents. \\
\texttt{target\_population} &
Population description used to sample personas. \\
\texttt{action\_space} &
Compact set of mutually interpretable actions used for aggregation and
comparison. \\
\texttt{survey\_questions} &
Optional structured questions asked of agents after or during decision making. \\
\texttt{network} &
Topology and neighbor sampling settings controlling social exposure. \\
\texttt{interaction\_protocol} &
Round-level exposure mechanism, such as independent survey or social diffusion. \\
\texttt{model\_mix} &
Ratio-controlled slots for LLM, VLM, custom-endpoint, random, and classical
agents. \\
\texttt{abm\_parameters} &
Parameters for selected classical rule families. \\
\texttt{comparison\_plan} &
Matched classical reference dynamics to launch with the current run. \\
\texttt{recording\_policy} &
Fields to preserve in prompts, decisions, costs, memories, and diagnostics. \\
\bottomrule
\end{tabular}
\caption{Core fields in the resolved AgoraSim configuration.}
\label{tab:config_schema}
\end{table*}

The configuration is intentionally compact. It is not meant to expose every
internal runtime object, but to preserve the assumptions that determine a
scenario trajectory: what agents perceive, what actions they can take, who can
influence whom, which agent types are used, which classical references are
launched, and which records are saved for audit.

\section{Interaction Protocols}
\label{app:protocols}

AgoraSim separates agent behavior from the exposure structure that governs what
agents observe across rounds. This separation matters because many scenario
trajectories depend not only on how an individual agent interprets an artifact,
but also on whether reactions are private, local, broadcast, sequential, or
experimentally assigned. The interaction protocol is therefore part of the
resolved configuration rather than a hidden platform assumption.
Table~\ref{tab:protocols} lists the protocols currently exposed by the demo.

\begin{table*}[t]
\centering
\small
\begin{tabular}{p{0.24\linewidth}p{0.34\linewidth}p{0.32\linewidth}}
\toprule
Protocol & Exposure rule & Typical use \\
\midrule
\texttt{independent\_survey} &
Agents receive the artifact and their persona context, but do not observe other
agents' statements. &
One-shot reaction testing, copy testing, survey-like simulation. \\
\texttt{social\_diffusion} &
Agents observe sampled public statements from network neighbors in the previous
round. &
Opinion spread, word of mouth, peer influence, adoption dynamics. \\
\texttt{broadcast\_comment\_stream} &
Agents observe a shared stream of salient or high-confidence public comments from
prior rounds. &
Public comment sections, viral posts, visible crowd response. \\
\texttt{sequential\_exposure} &
Agents receive staged messages or content variants across rounds. &
Campaign sequencing, clarifications, follow-up announcements, narrative shifts. \\
\texttt{focus\_group} &
Agents are assigned to small discussion groups with optional moderator framing. &
Deliberation, qualitative pretesting, local discussion dynamics. \\
\texttt{ab\_treatment} &
Agents are assigned to treatment labels or content variants under matched
settings. &
Message testing, framing comparisons, policy or product variants. \\
\bottomrule
\end{tabular}
\caption{Interaction protocols supported by AgoraSim. Each protocol controls how
agents encounter the artifact and one another while keeping the agent interface,
action space, and metrics fixed.}
\label{tab:protocols}
\end{table*}

All protocols share the same round interface. In round \(t\), each agent receives
the event, persona, action space, survey questions, optional media, prior memory,
and protocol-specific exposure inputs derived from earlier rounds. After agents
emit structured decisions, the environment stores public statements, updates
aggregate state, and prepares the exposure inputs for the next round. The
protocols are deliberately simple: they expose the main social-exposure
assumption behind a run rather than attempting to reproduce full platform ranking
or recommendation systems.

\section{Classical Reference Dynamics}
\label{app:classic}

Classical rules can be used in two ways: as \texttt{classic} agents inside the
current hybrid population, or as matched all-classical reference runs. In both
cases, the rule output is projected into the same action and metric schema as
LLM and VLM agents. Table~\ref{tab:classic_params} summarizes the main editable
parameters exposed by the demo.

\begin{table*}[t]
\centering
\small
\begin{tabular}{p{0.22\linewidth}p{0.30\linewidth}p{0.38\linewidth}}
\toprule
Reference dynamic & Example parameters & Interpretation \\
\midrule
Threshold / Bass &
social influence \(\beta\), adoption threshold, Bass \(p\), Bass \(q\), noise &
Controls independent uptake, imitation pressure, and adoption difficulty. \\
Bounded confidence / Deffuant &
confidence bound, convergence rate, noise &
Controls whether agents move only toward sufficiently similar opinions. \\
SIR contagion &
transmission probability, recovery probability, initial active share &
Controls exposure-driven spread and decay of attention or alarm. \\
Herding &
herding strength, independent reconsideration rate, noise &
Controls how strongly agents follow visible crowd signals. \\
DeGroot / voter &
neighbor weight, majority-switching probability, stubbornness &
Controls consensus formation and peer-driven opinion movement. \\
Discrete choice &
risk, trust, cost sensitivity, habit, social proof weights &
Controls utility-based action selection under interpretable features. \\
\bottomrule
\end{tabular}
\caption{Editable parameters for classical reference dynamics.}
\label{tab:classic_params}
\end{table*}

The default values are demonstration defaults rather than calibrated
domain-specific theories. Users can edit parameters before launch, and the saved
configuration records those choices. The purpose of these dynamics is to provide
interpretable reference trajectories: a close match can suggest that a pattern is
explainable by a simple rule family, while a divergence can identify assumptions
that deserve inspection or empirical validation.

\section{SDK, CLI, and REST Usage}
\label{app:interfaces}

The Python SDK exposes the same compact experiment facade as the UI. The example
below launches a small congestion-pricing scenario and saves the resolved result
bundle.

\begin{CodeBlock}
from agorasim.experiments import run_experiment

run = run_experiment(
    brief="congestion pricing",
    target_population="urban commuters",
    question="support, oppose, or adapt?",
    budget_usd=0.50,
    n_agents=80,
    rounds=3,
    model_policy="latest_within_budget",
)
run.to_json("results/congestion_pricing.json")
\end{CodeBlock}

The CLI accepts JSON or YAML experiment files and writes the resolved
configuration and result bundle. This supports batch runs, sensitivity sweeps,
and repository-based experiments where content variants, model mixes,
interaction protocols, seeds, and ABM reference strategies are varied
systematically.

The REST API exposes the same facade through
\texttt{POST /api/experiments/simple}. A typical request includes the scenario
brief, target population, question, budget, number of agents, rounds, model
policy, optional media metadata, and overrides. The response includes an
experiment id, status, resolved configuration, estimated cost, and comparison
run ids. Follow-up endpoints return results, records, reports, and saved
configurations.

\end{document}